\documentclass[conference]{IEEEtran}
\usepackage{cite}
\usepackage{amsmath,amssymb,amsfonts}

\usepackage{graphicx}
\usepackage{textcomp}
\usepackage{xcolor}
\usepackage{titlesec}
\usepackage{booktabs}
\usepackage{tabularx}
\usepackage{fancyhdr}
\usepackage{etoolbox} 
\def\BibTeX{{\rm B\kern-.05em{\sc i\kern-.025em b}\kern-.08em
    T\kern-.1667em\lower.7ex\hbox{E}\kern-.125emX}}
\usepackage{algorithm}
\usepackage{algpseudocode}

\begin{document}

\title{A U-Net and Transformer Pipeline for Multilingual Image Translation\\
}

\author{\IEEEauthorblockN{Siddharth Sahay}
\IEEEauthorblockA{\textit{Department of Machine Learning} \\
\textit{B.M.S. College of Engineering}\\
Bangalore, India \\
siddharthsahay2004@gmail.com}
\and
\IEEEauthorblockN{Radhika Agarwal}
\IEEEauthorblockA{\textit{Department of Machine Learning} \\
\textit{B.M.S. College of Engineering}\\
Bangalore, India \\
aradh2212@gmail.com}
}

\maketitle

\begin{abstract}
This paper presents an end-to-end multilingual translation pipeline that integrates a custom U-Net for text detection, the Tesseract engine for text recognition, and a from-scratch sequence-to-sequence (Seq2Seq) Transformer for Neural Machine Translation (NMT). Our approach first utilizes a U-Net model, trained on a synthetic dataset , to accurately segment and detect text regions from an image. These detected regions are then processed by Tesseract to extract the source text. This extracted text is fed into a custom Transformer model trained from scratch on a multilingual parallel corpus spanning 5 languages. Unlike systems reliant on monolithic pre-trained models, our architecture emphasizes full customization and adaptability. The system is evaluated on its text detection accuracy, text recognition quality, and translation performance via BLEU scores. The complete pipeline demonstrates promising results, validating the viability of a custom-built system for translating text directly from images.
\end{abstract}


\section{Introduction}
The ability to accurately interpret and translate multilingual content is essential across domains, from education \cite{b6} to global travel. This paper presents a fully custom-built pipeline capable of translating text across multiple languages directly from document images.

Traditionally, translators with OCR often have issues with script generalization and noise sensitivity, particularly in low-resource scenarios \cite{b5}. Our purpose is to address these challenges by proposing an end-to-end architecture that first isolates text regions, then recognizes the text, and finally translates it. This architecture integrates a U-Net-based text detection module \cite{b9} with the Tesseract recognition engine, followed by a Transformer-based sequence-to-sequence translation model \cite{b8}. A primary goal of this work is to build this entire pipeline with minimal reliance on external pre-trained weights.

Our pipeline's first stage employs a U-Net \cite{b9}, leveraging its strong spatial encoding and localization capabilities originally designed for biomedical segmentation. This U-Net model is trained to act as a highly effective text segmentation module. It processes the input image and produces a binary mask, precisely identifying where text is located. This segmentation step effectively isolates text from confusing backgrounds, fonts, and multilingual scripts.

The text regions identified by the U-Net are then cropped and fed into the Tesseract OCR engine. The recognized text extracted by Tesseract is then passed to our custom multilingual Transformer model. This NMT model was trained from scratch on a multilingual parallel corpus \cite{b10}, \cite{b11} spanning 5 languages and a dataset having 2.2 million translation pairs.

The uniqueness of this system lies not only in its holistic design but also in its independence from large-scale models, such as Google's NMT system \cite{b2}. This design choice enables full control over language tokenization, vocabulary granularity, and performance tuning. We aim to demonstrate that an effective document translation system is achievable without total reliance on massive datasets, highlighting the value of an adaptable and efficient architecture. This system could help aid in education \cite{b6} or assist tourists who need to translate written text like road signs into a language they understand. Our test results validate the effectiveness of this integrated system, demonstrating strong performance in both recognition accuracy and translation quality.

\section{Literature Survey}
This section reviews key advancements in multilingual NMT, Transformer architectures, and the integration of OCR with translation systems, identifying the research gaps our work aims to address.
\subsection{Advancements in Multilingual Neural Machine Translation}
The field of Neural Machine Translation (NMT) has seen significant progress in recent years \cite{b1}, enabling large-scale multilingual systems. Meta AI's M2M-100 model, for instance, can translate between 100 languages without relying on English as an intermediate (or "pivot") language. Similarly, Google's Multilingual NMT System (GNMT) \cite{b2} employs deep LSTM models with attention to achieve high-quality zero-shot translations. While these models demonstrate impressive capabilities, their development requires extensive computational resources and massive-scale datasets, posing a significant barrier for custom or resource-constrained applications

\subsection{Transformer-Based Translation Models}
The Transformer architecture \cite{b8} has become the foundation of modern NMT due to its superior handling of long-range dependencies through self-attention mechanisms. Its efficacy has been validated in numerous studies, such as a French-to-English translation model that achieved high accuracy using a standard Transformer network \cite{b3}. Researchers have also explored enhancements, such as bilingual attention mechanisms that incorporate decoded history to improve translation quality across various language pairs \cite{b4}. However, these models still demand substantial parallel corpora and significant computational power for training.

\subsection{Integration of OCR and NMT Systems}
The fusion of Optical Character Recognition (OCR) with NMT is critical for translating text from images and scanned documents. Most existing systems, however, treat OCR and translation as separate, disconnected modules, which can lead to error propagation. The seamless integration of these components into a robust pipeline remains a challenging area, especially for documents containing multiple languages or complex scripts.

\subsection{Application in Education and Real-Time Communication}
The practical applications for integrated NMT systems are growing. In educational contexts, NMT is being explored as a tool to aid in foreign language teaching and learning \cite{b6}. Concurrently, the demand for integrated language tools is highlighted by the development of AI-powered real-time text editors that feature multilingual translation and speech recognition capabilities \cite{b7}.

\subsection{Research Gaps}
Our review of the existing literature reveals several key gaps:

\subsubsection{Integrated Detection-Recognition-Translation Systems}
There is a lack of systems that seamlessly integrate the full pipeline, from initial text detection to recognition and finally translation. Many approaches fail to optimize the handoff between these stages, particularly for handling noisy images with diverse, multi-script text.

\subsubsection{Resource Efficiency}
As noted, current state-of-the-art models \cite{b2} require computational resources that are inaccessible for many custom applications. This creates a need for effective, from-scratch models that can be trained and deployed more efficiently.

\subsubsection{Low-Resource Language Support}:
Despite progress, effectively supporting low-resource languages remains a significant hurdle in both OCR \cite{b5} and NMT.

Our work addresses these gaps by proposing a fully integrated, end-to-end pipeline. We specifically tackle Gap 1 by using a U-Net \cite{b9} as a dedicated text detector to pre-process images for a standard recognition engine, improving robustness. We address Gap 2 by building and training our own Transformer model from scratch, demonstrating a viable, customizable alternative to massive pre-trained models

\section{Proposed Method}
This section details the procedures and architectures used to build the complete translation system. The methodology is broken into two parts: data collection and the multi-stage model pipeline.

\subsection{Data Collection and Preparation}
Building a robust model requires appropriate training data for each component.

\subsubsection{Text Detection (U-Net) Dataset}
To train the text detection module, we required images with corresponding word-level masks. We generated a synthetic dataset of text images with varied backgrounds, orientations, fonts, and font sizes . This was accomplished using a Python program that rendered words onto background images and saved both the final image and its corresponding binary mask (text in white, background in black). The synthetic data generation process is detailed in Algorithm~\ref{alg:ocr_synthetic_generation}.

\begin{algorithm}[h!]
\caption{Synthetic Data Generation for Multilingual OCR}
\label{alg:ocr_synthetic_generation}
\begin{algorithmic}[1]
\State \textbf{Input:} Number of samples $N$, font directory $\mathcal{F}$, word lists $\mathcal{W}$, output directory $\mathcal{O}$
\State \textbf{Output:} Rendered multilingual text images and ground truth labels

\State Load multilingual fonts from $\mathcal{F}$ for each language (e.g., English, French, German, Russian, Italian)
\State Load word lists $\mathcal{W}_\text{lang}$ for each supported language
\For{$i = 1$ to $N$}
    \State Randomly select a language $\ell$ from the available languages
    \State Randomly sample up to 5 words from $\mathcal{W}_\ell$
    \State Concatenate sampled words to form a text string $T$
    \State Retrieve the corresponding font $\mathcal{F}_\ell$ and set font size
    \State Compute text dimensions using the selected font
    \State Create a blank grayscale image canvas of the same size
    \State Render text $T$ onto the canvas using the chosen font
    \State Save rendered image as \texttt{synthetic\_text\_\{$i$\}\_\{$\ell$\}.png} in $\mathcal{O}$
    \State Save the corresponding text label $T$ in a \texttt{.txt} file with the same name
\EndFor
\State \textbf{Return:} A set of image-text pairs for OCR model training
\end{algorithmic}
\end{algorithm}

\subsubsection{Translator (NMT) Dataset}
For the translation model, parallel corpora were collected for Russian, Italian, German, and French, sourced from online repositories like Tatoeba \cite{b10} and OPUS-100 \cite{b11}. Each language was paired with English, creating translation files in both directions (e.g., en-fr and fr-en). This setup, totaling 2.2 million translation pairs, allows the model to learn translations not only to and from English but also between the non-English languages (e.g., German to French). The text preprocessing and tokenization procedure is described in \textbf{Algorithm~\ref{alg:text_preprocessing}}.

\begin{algorithm}[h!]
\caption{Text Preprocessing for Multilingual Transformer}
\label{alg:text_preprocessing}
\begin{algorithmic}[1]
\State \textbf{Input:} CSV file with source/target texts and language codes
\State \textbf{Output:} Tokenized and numericalized source-target sentence pairs
\State Load CSV and extract columns: \texttt{source\_text}, \texttt{target\_text}, \texttt{source\_language}, \texttt{target\_language}
\For{each row in the dataset}
    \State Tokenize source text as: \texttt{<src\_lang>} + tokens + \texttt{<tgt\_lang>}
    \State Tokenize target text as: \texttt{<tgt\_lang>} + tokens + \texttt{<eos>}
\EndFor
\If{building vocabulary}
    \State Construct source vocabulary from all source tokens
    \State Construct target vocabulary from all target tokens
    \State Include special tokens: \texttt{<pad>}, \texttt{<sos>}, \texttt{<eos>}, \texttt{<unk>}
\EndIf
\For{each tokenized sentence}
    \State Add \texttt{<sos>} and \texttt{<eos>} tokens to source sentence
    \State Add \texttt{<sos>} token to target sentence
    \State Convert tokens to indices using vocabulary
\EndFor
\State Define batching procedure with padding:
\begin{itemize}
    \item Pad source and target sequences to the max length in batch using \texttt{<pad>} token
\end{itemize}
\State Return tokenized, indexed, and padded dataset
\end{algorithmic}
\end{algorithm}

\subsection{System Architecture and Model Development}
Our system is a three-stage pipeline: (1) Text Detection, (2) Text Recognition, and (3) Text Translation.

\subsubsection{U-Net for Text Detection}
The first stage is a U-Net model \cite{b9} designed for image segmentation. Its purpose is to take an input image and produce a binary mask that precisely identifies all word-level text regions.

\begin{itemize}
    \item \textbf{Architecture:} The model uses a standard U-Net encoder-decoder framework. The encoder consists of convolutional blocks (two conv-layers with ReLU, followed by max pooling) for downsampling. The decoder mirrors this structure, using transposed convolutions for upsampling. Skip connections concatenate encoder-layer feature maps with corresponding decoder layers, preserving spatial information crucial for accurate mask localization. The final layer uses a sigmoid activation to output the binary mask.
    \item \textbf{Model Configuration:} Hyperparameters are summarized in \textbf{Table~\ref{tab:unet-config}}. Input images are resized to $512 \times 512$, and input dimensions are multiples of 16 to prevent shape mismatches.
    \item \textbf{Training Setup:} The model was trained using the Adam optimizer with a learning rate of $1 \times 10^{-4}$ 12and a binary cross-entropy loss function. We used a batch size of 8 and reserved 20\% of the data for validation.
    \item \textbf{Data Augmentation:} To improve generalization, we applied random rotations, flips, color jittering, and elastic deformations.
\end{itemize}

\begin{table}[h]
\centering
\caption{U-Net Model Hyperparameters}
\label{tab:unet-config}
\begin{tabular}{|l|c|}
\hline
\textbf{Parameter} & \textbf{Value} \\
\hline
Input image size & 512 $\times$ 512 $\times$ 3 \\
Encoder depth & 4 levels \\
Filter size & 3 $\times$ 3 \\
Activation & ReLU \\
Batch normalization & Yes \\
Dropout & 0.3 \\
\hline
\end{tabular}
\end{table}

\subsubsection{Text Region Cropping and Recognition}
The binary mask from the U-Net is a pixel-level output and cannot be directly fed to a recognition engine. To bridge this gap, we first apply a contour-finding algorithm to the mask. This process identifies the outer boundaries of each connected white region (i.e., each detected word). From these contours, we compute a bounding box for each word.

These bounding boxes are then used to crop the corresponding regions from the original input image. Each cropped image, now containing just one word, is passed to the Tesseract OCR engine. Tesseract performs the actual text recognition, converting the word image into a machine-readable string. This U-Net pre-processing step serves to isolate words from noisy backgrounds, simplifying the task for Tesseract.

\subsubsection{Transformer for Neural Machine Translation}
The final stage uses a sequence-to-sequence (Seq2Seq) Transformer model, based on the architecture from Vaswani et al. \cite{b8}, to perform the translation. This model was built from scratch, allowing full control over its architecture and training.

\begin{itemize}
    \item \textbf{Architecture:} The model follows the standard encoder-decoder framework. Both encoder and decoder use an embedding dimension of 512, with sinusoidal positional encodings added to preserve sequence information \cite{b8}. The encoder and decoder are each 6 layers deep, with 8 attention heads and a feed-forward network size of 2048.
    \item \textbf{Model Configuration:} Key hyperparameters are detailed in \textbf{Table~\ref{tab:transformer_params}}.
    \item \textbf{Training Setup:} We used the Adam optimizer with a fixed learning rate of $1 \times 10^{-4}$ 22 and a cross-entropy loss function23. The model was trained with a batch size of 64 for 5 epochs.
    \item \textbf{Vocabulary and Tokenization:} We used a dual vocabulary system (one for source, one for target) built from the training corpus. This includes special tokens ($<$sos$>$, $<$eos$>$, $<$pad$>$, $<$unk$>$) and ISO-639-1 language codes to specify the source and target languages (see \textbf{Algorithm~\ref{alg:text_preprocessing}}). Tokenization was performed using a simple whitespace-based approach, which is noted as a key area for future improvement.
    
    \item \textbf{Computation Consideration:} The model has approximately 60 million trainable parameters and was trained on an Nvidia RTX A5000 GPU.
\end{itemize}

\begin{table}[h]
\centering
\caption{Transformer Model Hyperparameters}
\begin{tabularx}{\columnwidth}{l X}
\toprule
\textbf{Parameter} & \textbf{Value} \\
\midrule
Hidden dimension ($d_{\text{model}}$) & 512 \\
Number of layers (encoder/decoder) & 6 \\
Attention heads & 8 \\
Feed-forward size & 2048 \\
Dropout rate & 0.3 \\
Max sequence length & 5000 \\
Attention head size & 64 \\
\bottomrule
\end{tabularx}
\label{tab:transformer_params}
\end{table}

\section{Results and Discussion}
This section presents the performance evaluation of our two-stage pipeline. We first analyze the U-Net text detection model and its integration with Tesseract, followed by an evaluation of the from-scratch NMT model.

\subsection{Text Detection and Recognition (OCR Pipeline}
\subsubsection{U-Net Model Convergence}

\begin{figure}[h]
    \centering
    \includegraphics[width=\columnwidth]{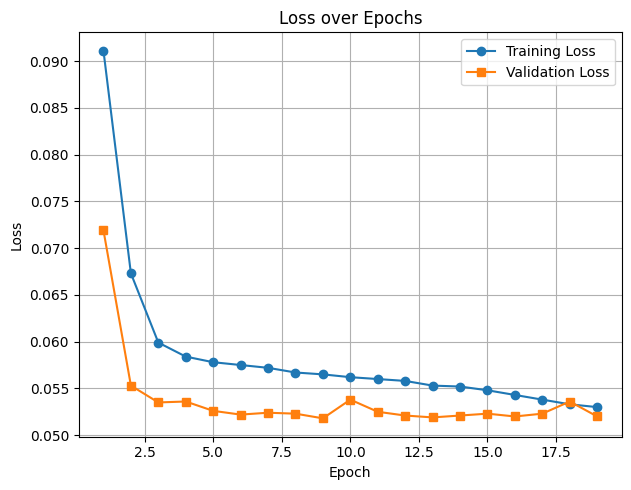}
    \caption{Loss Curve of the U-NET Model.}
    \label{fig:UNET loss}
\end{figure}

The training and validation loss for the U-Net text detection model is shown in \textbf{Fig.~\ref{fig:UNET loss}}. The model converges effectively, with the training loss decreasing from 0.0911 to 0.0530 and the validation loss stabilizing around 0.0520. The close alignment between the training and validation curves, with no significant divergence, suggests that the model is not overfitting and generalizes well to unseen data.

\subsubsection{Text extraction}
To extract textual information from the output of UNET model, an open-source Oprical Character Recognition (OCR) engine called Tesseract was used. Tesseract is well known for its robustness and accuravy in recognizing printed and hand written text acorss many different languages. In this project, it was integrated into the pipeling to convert the text detection output into machine readable text. 

\subsubsection{Implementation}
\begin{figure}[h]
    \centering
    \includegraphics[width=\columnwidth]{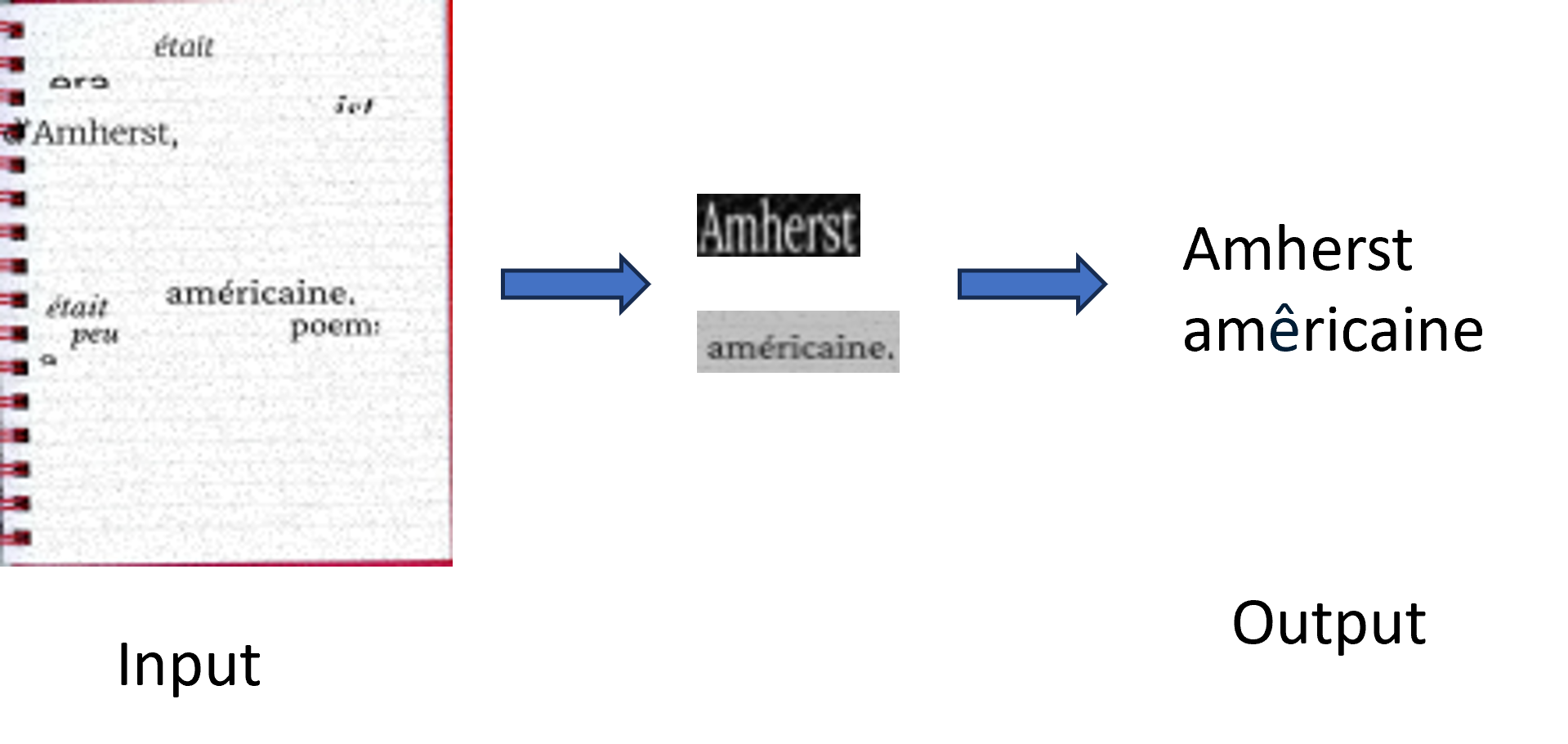}
    \caption{Text detection and extraction.}
    \label{fig:Workflow}
\end{figure}

\textbf{Fig.~\ref{fig:Workflow}} illustrates the complete detection and recognition pipeline. The input image (left) is processed by the U-Net, which detects the text regions. These regions are then cropped (center) and passed to Tesseract, which extracts the final text output (right)

\subsection{Multilingual Translation (NMT Model)}
\subsubsection{Translator Model Convergence}
\begin{figure}[h]
    \centering
    \includegraphics[width=\columnwidth]{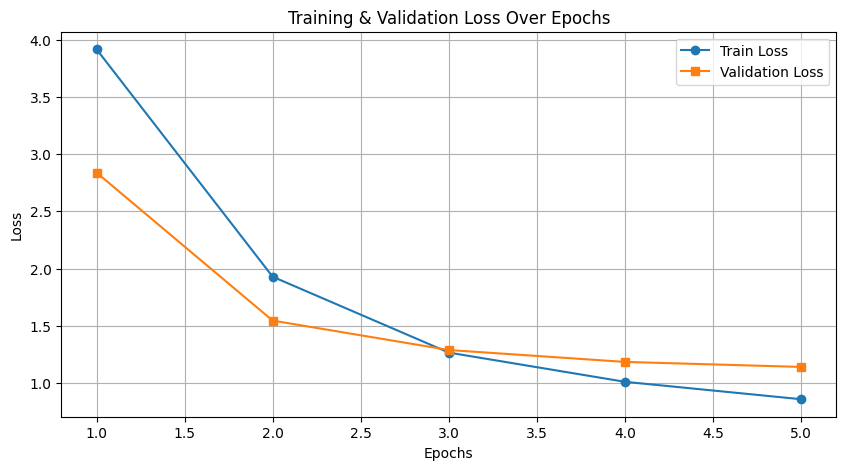}
    \caption{Loss Curve of the translator model.}
    \label{fig:translator loss}
\end{figure}

The loss curve for the multilingual Transformer model is shown in \textbf{Fig.~\ref{fig:translator loss}}. The model was trained for 5 epochs. The training loss converged to 0.8585, while the validation loss reached a minimum of 1.1405. The steady decrease in both losses demonstrates stable learning.

\subsubsection{Effects of Training Data volume}
To understand the impact of dataset size, we trained the model progressively with more data. \textbf{Table~\ref{tab:data_size_loss}} shows a clear trend: as the number of sentence pairs per language increases from 10,000 to 400,000, the minimum validation loss consistently decreases from 6.0935 to 1.1405. This highlights the model's scalability and confirms that its performance is heavily dependent on data volume.

\begin{table}[h]
\centering
\caption{Validation Loss vs. Data Size (per language)}
\begin{tabular}{l c}
\toprule
\textbf{Data Size (per language)} & \textbf{Minimum Validation Loss} \\
\midrule
10,000    & 6.0935 \\
50,000    & 4.5195 \\
70,000    & 3.6053 \\
100,000   & 2.7743 \\
200,000   & 2.1031 \\
300,000   & 1.5023 \\
400,000   & 1.1405 \\
\bottomrule
\end{tabular}
\label{tab:data_size_loss}
\end{table}

\subsubsection{Translation Quality Metrics}
We evaluated the final model on a held-out test set using standard machine translation metrics: BLEU \cite{b12}, METEOR \cite{b13}, ROUGE-L \cite{b14}, and TER \cite{b15}. The results are summarized in \textbf{Table~\ref{tab:translation_metrics}}.

\begin{table}[h!]
\centering
\caption{Translation Quality Evaluation Metrics}
\label{tab:translation_metrics}
\begin{tabular}{l c}
\toprule
\textbf{Metric} & \textbf{Score} \\
\midrule
BLEU    & 0.3168 \\
BLEU-1  & 0.6346 \\
BLEU-2  & 0.4818 \\
BLEU-3  & 0.3736 \\
BLEU-4  & 0.2807 \\
METEOR  & 0.6907 \\
ROUGE-L & 0.6527 \\
TER     & 0.5111 \\
\bottomrule
\end{tabular}
\end{table}

The model achieved a BLEU score of 0.3168, suggesting a moderate level of translation quality. The high BLEU-1 score (0.6346) indicates good word-level (unigram) accuracy. However, the lower BLEU-4 score (0.2807) suggests the model has difficulty capturing longer, more fluent phrase structures. This is an expected limitation of our simple whitespace-based tokenization, which struggles with compound words or different morphological forms.

The METEOR score of 0.6907 is promising, as it indicates the model effectively captures semantic meaning and synonyms, addressing some of BLEU's limitations. The ROUGE-L score (0.6527) further supports this, showing good structural similarity with reference translations. Finally, the TER (Translation Edit Rate) of 0.5111 means that approximately 51\% of the output would require editing to match the reference, which is a solid result for a model trained from scratch.

\subsubsection{Sample Translations:}

\textbf{Table~\ref{tab:translation_outputs}} shows sample translations produced by the model.

\begin{table}[t!]
\centering
\caption{Sample Translation Outputs from the Model}
\label{tab:translation_outputs}
\resizebox{\columnwidth}{!}{%
\begin{tabular}{llp{3.2cm}p{3.2cm}p{3.2cm}}
\toprule
\textbf{Source Lang} & \textbf{Target Lang} & \textbf{Source Text} & \textbf{Reference Translation} & \textbf{Model Hypothesis} \\
\midrule
English & French & I like to read books. & J'aime lire des livres. & jame lire des livres \\
English & French & I don't understand. & Je ne comprends pas. & je ne comprends pas \\
German  & English & Ich habe meine Kreditkarte verloren. & I've lost my credit card. & i lost my credit card. \\
English & Italian & It was only a hypothesis. & Era solamente un'ipotesi. & era soltanto un follow di \\
\bottomrule
\end{tabular}
}
\end{table}

The samples show the model is often correct (e.g., "je ne comprends pas") but also makes errors in word choice (e.g., "jame" instead of "J'aime") or coherence ("era soltanto un follow di"). This aligns with the quantitative metrics, showing good word-level understanding but weaknesses in fluency.

\section{Conclusion and Future Scope}

\subsection{Conclusion}

This work presented an end-to-end pipeline for image-based text translation, successfully integrating a U-Net for text detection, Tesseract for text recognition, and a custom-built Transformer for multilingual translation.

The U-Net text detection model demonstrated stable convergence with minimal overfitting, as indicated by the close alignment of its training and validation loss curves. We confirmed that the Tesseract OCR engine can be effectively integrated into this pipeline, reliably extracting text from the image regions identified by the U-Net.

Our from-scratch multilingual translation model achieved promising results across standard evaluation metrics. A \textbf{BLEU} score of \textbf{0.3168} and a \textbf{METEOR} score of \textbf{0.6907} suggest the model is capable of producing semantically meaningful and contextually relevant translations. Furthermore, the analysis of training data volume clearly emphasizes the significance of data size in improving model generalization and reducing validation loss. This project validates the feasibility of building a custom, adaptable, and end-to-end system for document translation.

\subsection{Future Scope}

Although the current system performs well, several areas offer potential for enhancement:

\begin{itemize}
\item \textbf{Improving Translation Fluency:} While the BLEU-1 score is high (0.6346), the lower BLEU-4 (0.2807) and higher TER (0.5111) scores indicate a need for improvement in generating fluent, coherent phrases . Implementing \textbf{subword tokenization} (like BPE or WordPiece) instead of simple whitespace splitting is a critical next step. Exploring advanced Transformer architectures such as \textbf{mBART} or \textbf{M2M-100} could also yield significant gains in fluency.

\item \textbf{OCR Accuracy Enhancement:} The current pipeline relies on Tesseract for recognition. Replacing or augmenting Tesseract with modern, deep learning-based OCR systems like \textbf{TrOCR} or \textbf{PaddleOCR} may improve recognition accuracy, particularly for handwritten, stylized, or low-resolution text.

\item \textbf{Domain Adaptation:} The model could be fine-tuned on domain-specific datasets (e.g., legal, medical, or technical text) to significantly enhance its performance in specialized applications.

\item \textbf{Real-time Deployment:} Optimizing the pipeline for speed would enable deployment in mobile and embedded systems, facilitating real-time applications such as translation wearables or smart assistants.

\item \textbf{Low-resource Language Support:} The framework can be extended to support more low-resource languages by leveraging data augmentation, transfer learning, and advanced multilingual training techniques.
\end{itemize}

This system provides a strong foundation for multilingual text extraction and translation, offering multiple directions for future research and development.


\begin{thebibliography}{00}
\bibitem{b6} B. Klimova, K. Pikhart, L. Hincz, and D. Marek, "Neural Machine Translation in Foreign Language Teaching and Learning: A Systematic Review," \textit{Educ. Inf. Technol.}, vol. 28, pp. 663–682, 2023. [Accessed: 12-March-2025].

\bibitem{b5} D. Kulshreshtha, R. Singh, P. Agrawal, and A. Kumar, "Multilingual Contextual Adapters To Improve Custom Word Recognition In Low-resource Languages," \textit{arXiv preprint arXiv:2307.00759}, 2023. [Online]. Available: https://arxiv.org/abs/2307.00759. [Accessed: 12-March-2025].

\bibitem{b9} O. Ronneberger, P. Fischer, and T. Brox, "U-Net: Convolutional Networks for Biomedical Image Segmentation," in \textit{Proc. Med. Image Comput. Comput.-Assist. Intervent. (MICCAI)}, Munich, Germany, 2015, pp. 234–241. [Online]. Available: https://arxiv.org/abs/1505.04597. [Accessed: 15-March-2025].

\bibitem{b8} A. Vaswani, N. Shazeer, N. Parmar, J. Uszkoreit, L. Jones, A. N. Gomez, Ł. Kaiser, and I. Polosukhin, "Attention is all you need," in 
\textit{Proc. 31st Int. Conf. Neural Inf. Process. Syst. (NeurIPS)}, Long Beach, CA, USA, 2017, pp. 6000–6010. [Accessed: 14-March-2025].

\bibitem{b10} C. Kelly, “Tab-delimited Bilingual Sentence Pairs from the Tatoeba Project,” \textit{ManyThings.org}, Apr. 1, 2024. [Online]. Available: https://www.manythings.org/anki/. [Accessed: 18-March-2025].

\bibitem{b11} Helsinki-NLP, “OPUS-100: Multilingual Parallel Corpus,” \textit{Hugging Face Datasets}, 2020. [Online]. Available: https://huggingface.co/datasets/Helsinki-NLP/opus-100. [Accessed: 18-March-2025].

\bibitem{b2} M. Johnson, M. Schuster, Q. V. Le, M. Krikun, Y. Wu, Z. Chen, N. Thorat, F. Viégas, M. Wattenberg, G. Corrado, M. Hughes, and J. Dean, "Google's Multilingual Neural Machine Translation System: Enabling Zero-Shot Translation," \textit{Trans. Assoc. Comput. Linguist.}, vol. 5, pp. 339–351, 2017. [Accessed: 11-March-2025].

\bibitem{b1} H. Dandge, M. Lokhande, and V. Jadhao, "Multilingual Global Translation using Machine Learning," in \textit{Proc. 2023 Int. Conf. Innovative Data Commun. Technol. Appl. (ICIDCA)}, 2023. [Online]. Available: https://doi.org/10.1109/ICIDCA56705.2023.10100287. [Accessed: 11-March-2025].

\bibitem{b3} H. Zhang, Y. Liu, and X. Wang, "A French-to-English Machine Translation Model Using Transformer Network," \textit{Procedia Comput. Sci.}, vol. 199, pp. 377–384, 2022. [Accessed: 11-March-2025].

\bibitem{b4} L. Kang, X. Chen, W. Huang, and Y. Li, "Bilingual Attention Based Neural Machine Translation," \textit{Appl. Intell.}, vol. 53, pp. 4302–4315, 2023. [Accessed: 11-March-2025].

\bibitem{b7} A. Sharma and V. Patel, "AI-Powered Real-Time Text Editor with Multilingual Translation and Speech Recognition," in \textit{Proc. 2025 Int. Conf. Artif. Intell. Mach. Learn. (ICAIML)}, 2025. [Accessed: 14-March-2025].

\bibitem{b12} K. Papineni, S. Roukos, T. Ward, and W.-J. Zhu, "BLEU: a Method for Automatic Evaluation of Machine Translation," in \textit{Proc. 40th Annu. Meeting Assoc. Comput. Linguistics (ACL)}, Philadelphia, PA, USA, 2002, pp. 311–318. [Online]. Available: https://aclanthology.org/P02-1040/. [Accessed: 18-March-2025].

\bibitem{b13} S. Banerjee and A. Lavie, "METEOR: An Automatic Metric for MT Evaluation with Improved Correlation with Human Judgments," in \textit{Proc. ACL Workshop on Intrinsic and Extrinsic Evaluation Measures for Machine Translation and/or Summarization}, Ann Arbor, MI, USA, 2005, pp. 65–72. [Online]. Available: https://www.cs.cmu.edu/~alavie/METEOR/. [Accessed: 18-March-2025].

\bibitem{b14} C.-Y. Lin, "ROUGE: A Package for Automatic Evaluation of Summaries," in \textit{Proc. Workshop on Text Summarization Branches Out (WAS 2004)}, Barcelona, Spain, 2004, pp. 74–81. [Online]. Available: https://aclanthology.org/W04-1013/. [Accessed: 18-March-2025].

\bibitem{b15} M. Snover, B. Dorr, R. Schwartz, L. Micciulla, and J. Makhoul, "A Study of Translation Edit Rate with Targeted Human Annotation," in \textit{Proc. 7th Conf. Assoc. Mach. Transl. Am. (AMTA)}, Cambridge, MA, USA, 2006, pp. 223–231. [Online]. Available: http://www.cs.umd.edu/~snover/tercom/. [Accessed: 18-March-2025].



\end{thebibliography}
\end{document}